\documentclass[11pt, a4paper,  notitlepage]{extarticle}

\usepackage[left=3cm, right=3cm, top=3.5cm, bottom=3.5cm, bindingoffset=0cm]{geometry}
\usepackage[T2A]{fontenc}
\usepackage[utf8]{inputenc}
\usepackage{graphicx}
\usepackage{mathtools}
\usepackage[affil-it]{authblk}
\usepackage{url}
\usepackage{amsmath}
\usepackage{amssymb}
\usepackage{multicol}
\usepackage{float}
\usepackage[section]{placeins}
\usepackage{listings}
\usepackage{changepage}
\title{Forecasting of  Non-Stationary Sales Time Series Using Deep Learning}
\author{Bohdan M.  Pavlyshenko \\  Ivan Franko National University of Lviv, Ukraine \\
  b.pavlyshenko@gmail.com,  www.linkedin.com/in/bpavlyshenko/}
\begin{document}
\maketitle
\sloppy
\begin{abstract}
The paper describes  the deep learning approach for forecasting non-stationary time series with using time trend correction  in a neural network model.  
Along with the layers for predicting sales values, the neural network model includes a subnetwork block for the prediction weight for a time trend term which is added to a predicted sales value. The time trend term is considered as a product of the predicted weight value and normalized time value.  The results show that 
 the forecasting accuracy can be essentially improved for non-stationary sales with time trends using the trend correction block in the deep learning model.  

Keywords: Sales forecasting, non-stationary time series, deep learning, trend correction.
\end{abstract}
\section{Introduction}
Sales and demand forecasting are widely being used in business analytics~\cite{mentzer2004sales,efendigil2009decision,zhang2004neural}. 
Sales can be treated as time series.
Different time series approaches are described in~\cite{chatfield2000time,brockwell2002introduction,box2015time,doganis2006time,hyndman2018forecasting,
tsay2005analysis,wei2006time,cerqueira2018arbitrage,hyndman2007automatic,papacharalampous2017comparison,tyralis2017variable,
tyralis2018large, papacharalampous2018predictability, taieb2012review, graefe2014combining}. 
Machine learning is widely used  for forecasting different kinds of time series along with classical statistical methods like ARIMA,  Holt-Winters,  etc.  
 Moderm deep learning algorithms DeepAR~\cite{salinas2020deepar, alexandrov2020gluonts}, \mbox{N-BEATS}~\cite{oreshkin2019n}, Temporal Fusion Transformers~\cite{lim2019temporal} show state-of-the-art results for time series forecasting.  
 Sales prediction is more a regression problem than a time series problem. 
The use of regression approaches for sales forecasting can often give us  better results compared to time series methods. 
Machine-learning algorithms make it possible to find patterns in the time series.
Some of the most popular ones are  tree-based machine-learning algorithms~\cite{james2013introduction}, e.g., Random Forest~\cite{breiman2001random}, Gradiend Boosting Machine~\cite{friedman2001greedy, friedman2002stochastic}.  
 The important time series features for their successful forecasting  are 
  their stationarity and sufficiently long time of historical observation to be able to capture intrinsic time series patterns. 
 One of the main assumptions of regression methods is that the patterns in the past data will be repeated in future.  
 There are some limitations of time series approaches for sales forecasting.  Let us consider some of them. 
We need to have historical data for a long time period to capture seasonality. 
However, often we do not have historical data for  a target variable, for example in case when a new product is launched. 
At the same time,  we have sales time series for a similar product and we can expect that our new product will have  a similar  sales pattern. 
 Sales data can have  a lot of outliers and missing data. We must clean those outliers and interpolate data before using a time series approach.
 We need to take into account a lot of exogenous factors which impact on sales.   
 On the other hand,  sales time series  have their own specifics,  e.g. their dynamics is caused by rather exogenous factors than intrinsic patterns, they are often highly non-stationary,  we frequently face with short time sales observations, e.g.  in the cases when new products or stores are just launched. 
 Often non-stationarity is caused by a time trend.  Sales trends can be different for different stores, e.g. some stores can have an ascending trend,  while others have a descending one. 
  Applying machine learning regression to non-stationary data, bias in sales prediction  on validation dataset can appear. This bias can be corrected by additional linear regression on a validation dataset when a covariate stands for predicted sales and a target variable stands for  real sales~\cite{pavlyshenko2019machine}. 

 In~\cite{pavlyshenko2016linear}, we studied linear models, machine learning, and probabilistic models for time series modeling. 
For probabilistic modeling, we considered the use of copulas and Bayesian inference approaches. 
In~\cite{pavlyshenko2018using}, we studied stacking approaches for time series forecasting and logistic regression with highly imbalanced data. 
In~\cite{pavlyshenko2019machine}, we study the usage of machine-learning models for sales time series forecasting.  
In~\cite{pavlyshenko2020salests}, we analyse sales time series using Q-learning from the perspective of the  dynamic price and supply optimization.
In~\cite{pavlyshenko2020bayesian},  we study the Bayesian approach for stacking machine learning predictive models for sales time series forecasting. 

 In this case study,  we consider the deep learning approach for forecasting non-stationary time series with the trend using a trend correction block in the deep learning model.   
Along with the layers for predicting sales values, the model includes a  subnetwork block for the prediction weight for a trend term which is added to the predicted sales value.  The trend term is considered as a product of the predicted weight value and normalized time value. 
  
\section{Sales time series with time trend}

\begin{figure}[H]
\center
 \includegraphics[width=1\linewidth]{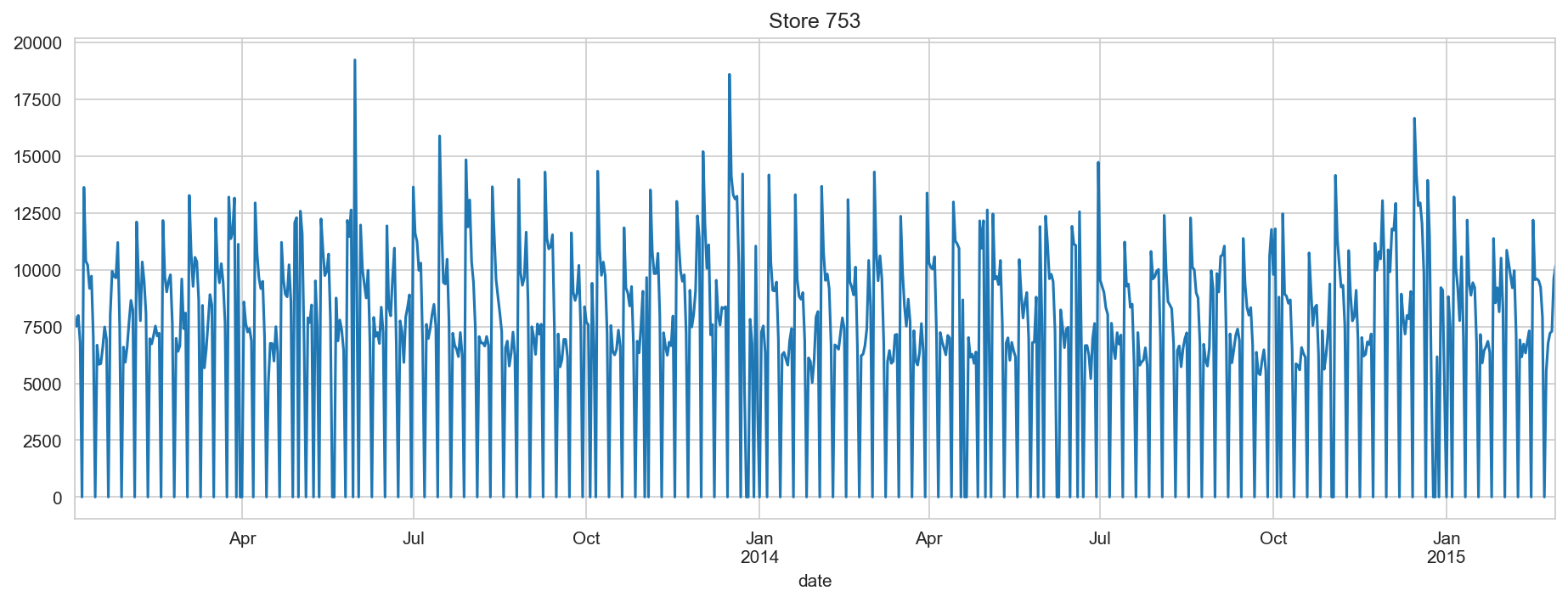}
 \caption{Store sales time series}
 \label{img1}
 \end{figure}

\begin{figure}[H]
\center
 \includegraphics[width=1\linewidth]{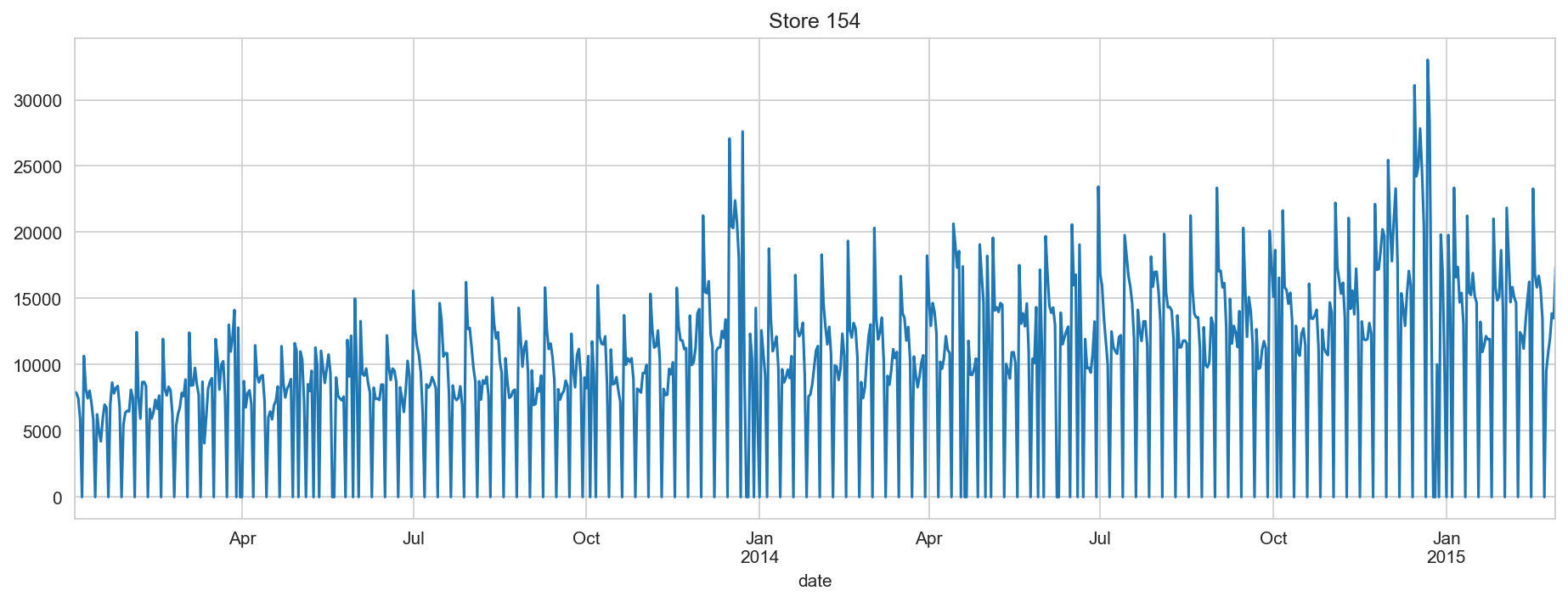}
 \caption{Store sales time series}
 \label{img2}
 \end{figure}

\begin{figure}[H]
\center
 \includegraphics[width=1\linewidth]{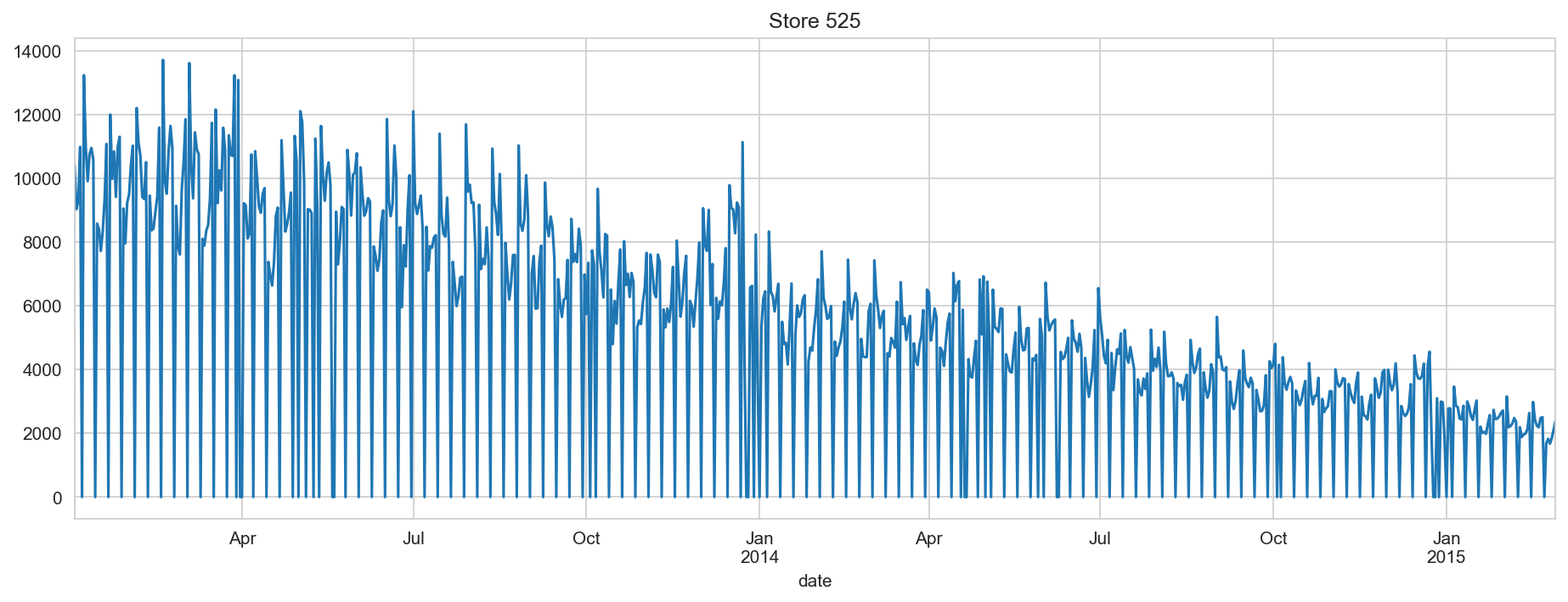}
 \caption{Store sales time series}
 \label{img3}
 \end{figure}

\begin{figure}[H]
\center
 \includegraphics[width=0.55\linewidth]{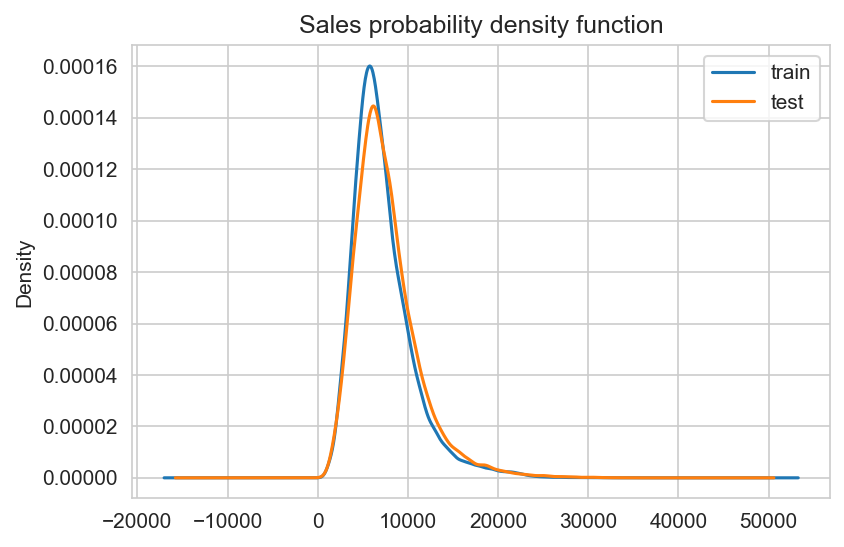}
 \caption{Probability density function for sales in all  stores}
 \label{img7}
 \end{figure}

\begin{figure}[H]
\center
 \includegraphics[width=0.55\linewidth]{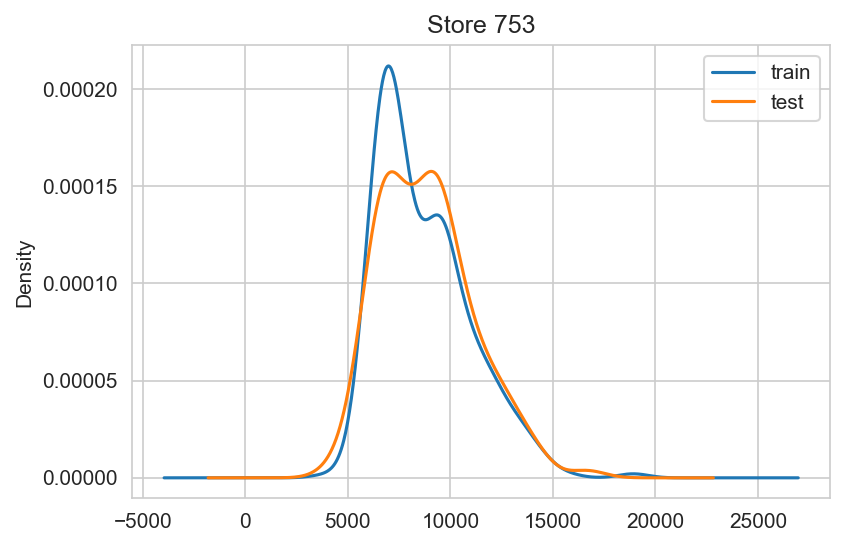}
 \caption{Probability density function for sales in specified  store}
 \label{img8}
 \end{figure}

\begin{figure}[H]
\center
 \includegraphics[width=0.55\linewidth]{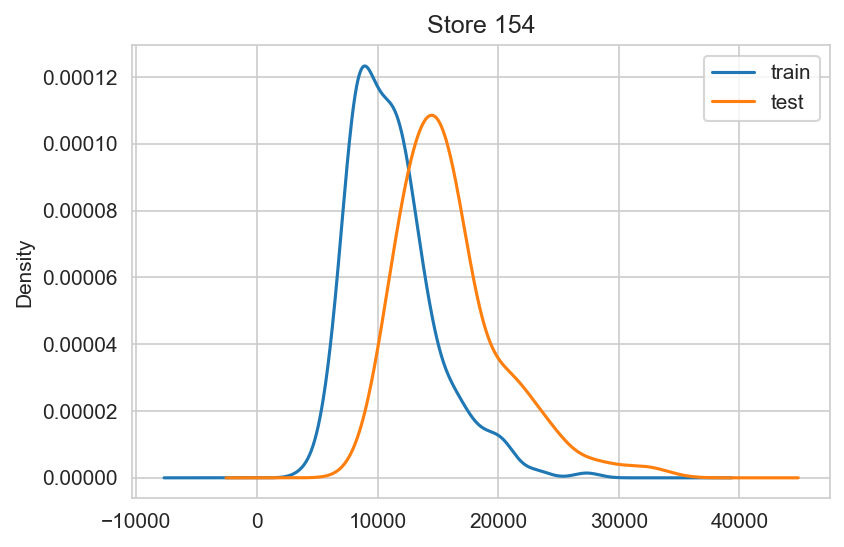}
 \caption{Probability density function for sales in specified  store}
 \label{img9}
 \end{figure}

\begin{figure}[H]
\center
 \includegraphics[width=0.55\linewidth]{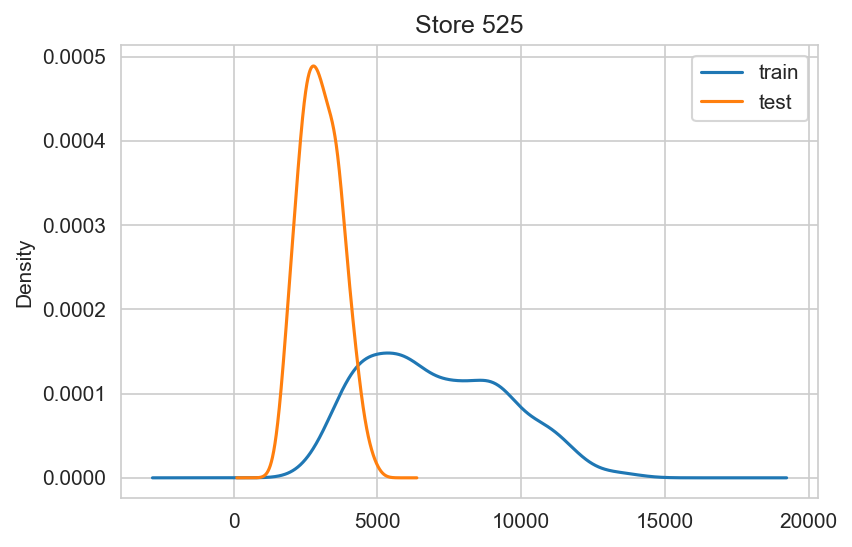}
 \caption{Probability density function for sales in specified  store}
 \label{img10}
 \end{figure}

  For the study, we have used the sales data which are based on the dataset from the 'Rossman Store Sales' Kaggle competition~\cite{rossmanstorekaggle}. 
  These data represent daily sales aggregated on the granularity level of customers and stores.  As the main features, \textit{ 'month','weekday',  'trendtype', 'Store', 'Customers',  'StoreType', 'StateHoliday','SchoolHoliday','CompetitionDistance'} were considered. 
  The \textit{'trendtype'} feature can be used in the case if the sales trend has different behavior types  on different time periods.  
  To study non-stationarity, arbitrary time trends were artificially added to the data grouped by stores.  
  The calculations were conducted in the Python environment using the main packages \textit{pandas, sklearn, numpy, keras, matplotlib, seaborn}.  To~conduct the analysis, \textit{Jupyter Notebook} was used. 
  Figures~\ref{img1}-\ref{img3} show the arbitrary examples of 
  aggregated sales time series with different  time trends for different stores. 
  Both training and validation datasets were received by splitting the dataset by date, so the validation time period is next with respect to the training period.  
Figure~\ref{img7} shows the probability density function (PDF) for all stores for training and validation datasets.  
  Figures~\ref{img8}--\ref{img10} show the probability density function for sales in data  samples, of specified stores which correspond to the stores time series shown in 
   Figures~\ref{img1}--\ref{img3}.
For some stores,  sales  PDF are different for training and validation datasets due to non-stationarity caused by different sales trends for different stores. 

\section{Deep learning model with time trend correction}
\begin{figure}[H]
\center
 \includegraphics[width=0.7\linewidth]{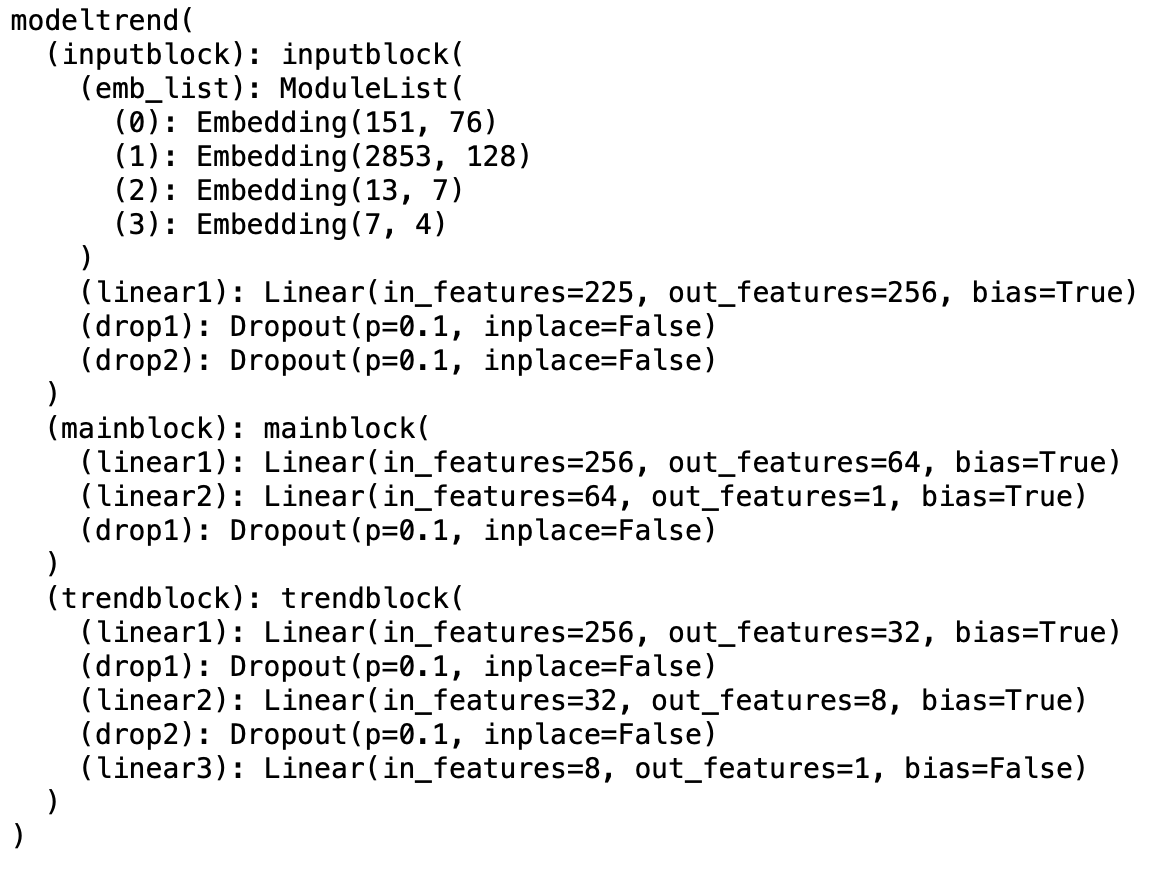}
 \caption{ Parameters of neural network layers}
 \label{model1}
 \end{figure}

\begin{figure}[H]
\center
 \includegraphics[width=0.8\linewidth]{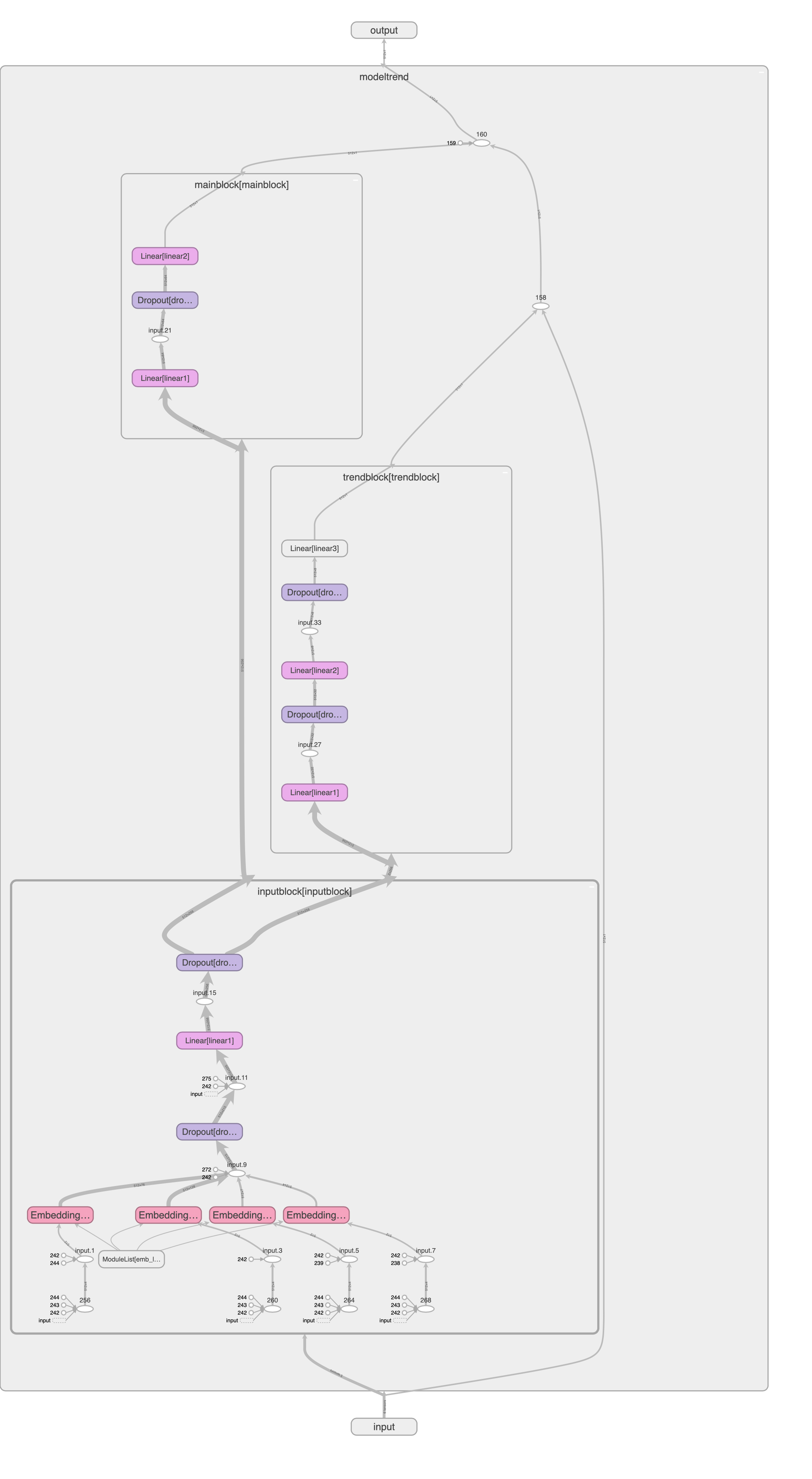}
 \caption{Neural network structure}
 \label{model2}
 \end{figure}

Let us consider  including a correction trend bock into the deep learning model. 
Along with the layers for predicting sales values, the model  will include a subnetwork block for the prediction weight for the time trend term which is added to the predicted sales value. The time trend term is considered as a product of the predicted weight value and normalized time value. 
The predicted sales values and the time trend term are combined in the loss function. As a result,  one can receive  an optimized trend correction for non-stationary sales for different groups of data with different trends.  
For modeling and deep learning case study,  the Pytorch deep learning library~\cite{paszke2017automatic, paszke2019pytorch}  was used. 
Categorical variables \textit{Store, Customer} with a large number of unique values were coded using embedding layers separately for each variable,  categorical variable with a small number of unique values \textit{StoreType,  Assortment} were represented using one-hot encoding. 
Figure~\ref{model1} shows the  parameters of neural network layers. 
Figure~\ref{model2} shows the neural network structure.

\begin{figure}[H]
\center
 \includegraphics[width=0.55\linewidth]{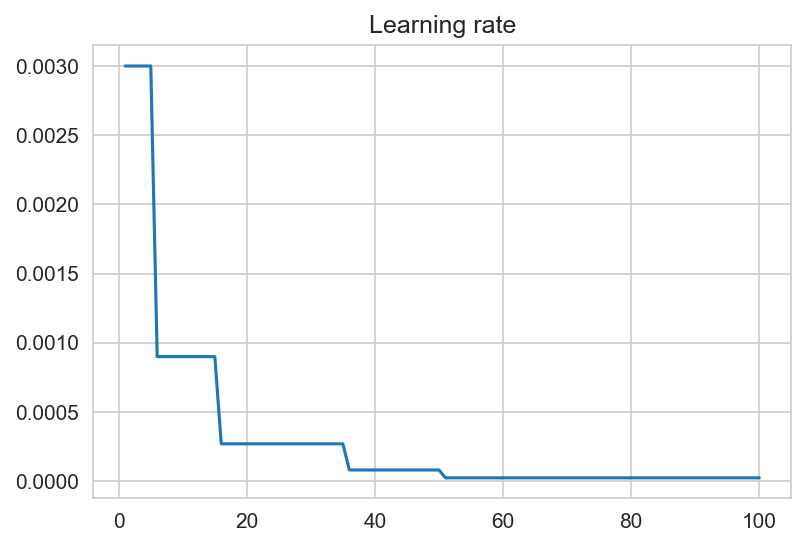}
 \caption{Learning rate changes with epochs}
 \label{img14}
 \end{figure}

\begin{figure}[H]
\center
 \includegraphics[width=0.55\linewidth]{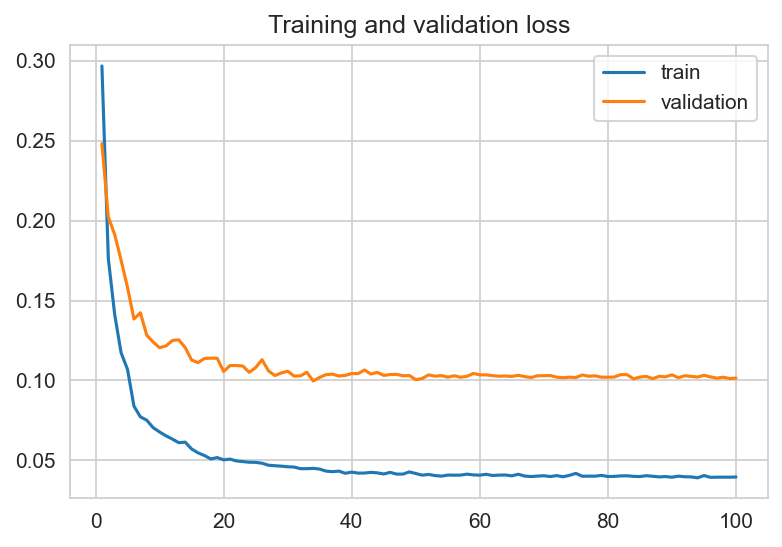}
 \caption{Train and validation loss  for model without trend correction block}
 \label{img15}
 \end{figure}

\begin{figure}[H]
\center
 \includegraphics[width=0.55\linewidth]{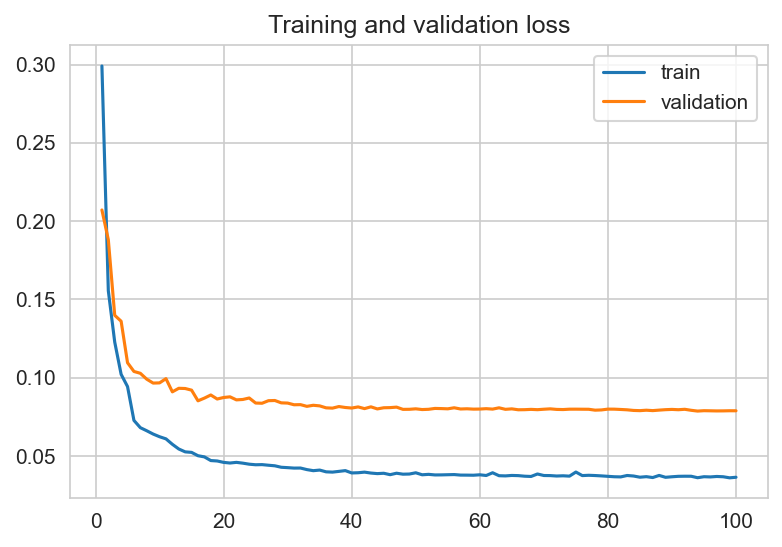}
 \caption{Train and validation loss for model with trend correction block}
 \label{img17}
 \end{figure}

\begin{figure}[H]
\center
 \includegraphics[width=0.65\linewidth]{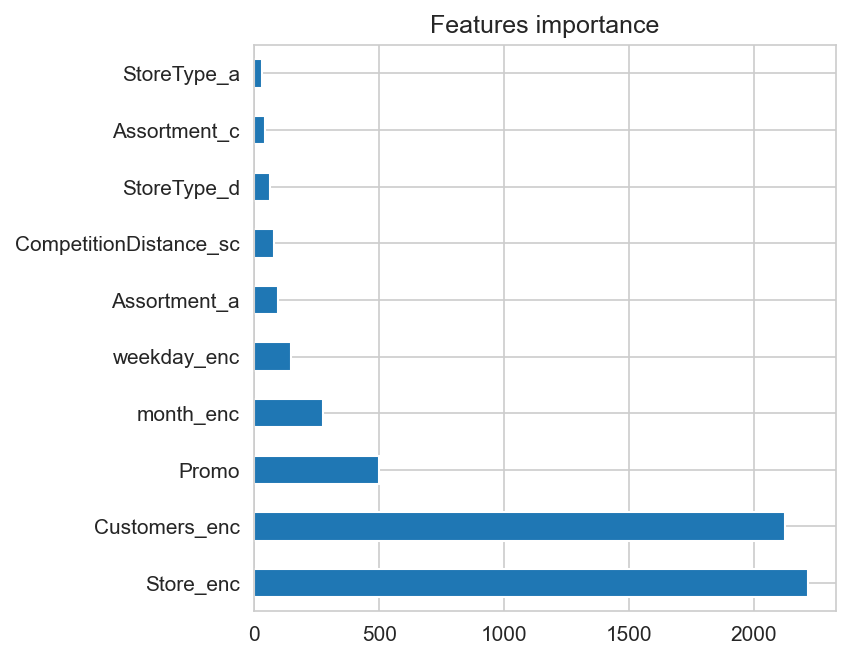}
 \caption{Features importance}
 \label{img18}
 \end{figure}

 The tensors of the output of embedding layers and numerical input values are concatenated 
and connected with the linear layer with \textit{ReLU} activation that forms an input block.  The output of this input block  is directed to the main block which consists of fully connected linear layers with \textit{ReLU} activation and dropout layers for predicting sales values using the output linear layer.  The output of the input block  is also directed to the trend correction block which consists of fully connected layer with ReLU activation,  dropout layer and output linear layer for the prediction of the trend weight.  The time trend term which  is a normalized time value multiplied by the trend weight is added to 
the predicted sales values.  For the comparison, we have considered two cases of the model with and without trend correction block.  Let us consider the results of model training and evaluation.  
Figure~\ref{img14} shows learning rate changes with epochs. 
Figure~\ref{img15} shows train and validation loss  for the model without trend correction block,  
Figure~\ref{img17} shows these losses  for the model with the trend correction block.
Figure~\ref{img18} shows the features importance which has been received  using the permutation approach. 
 
\begin{figure}[H]
\center
 \includegraphics[width=0.65\linewidth]{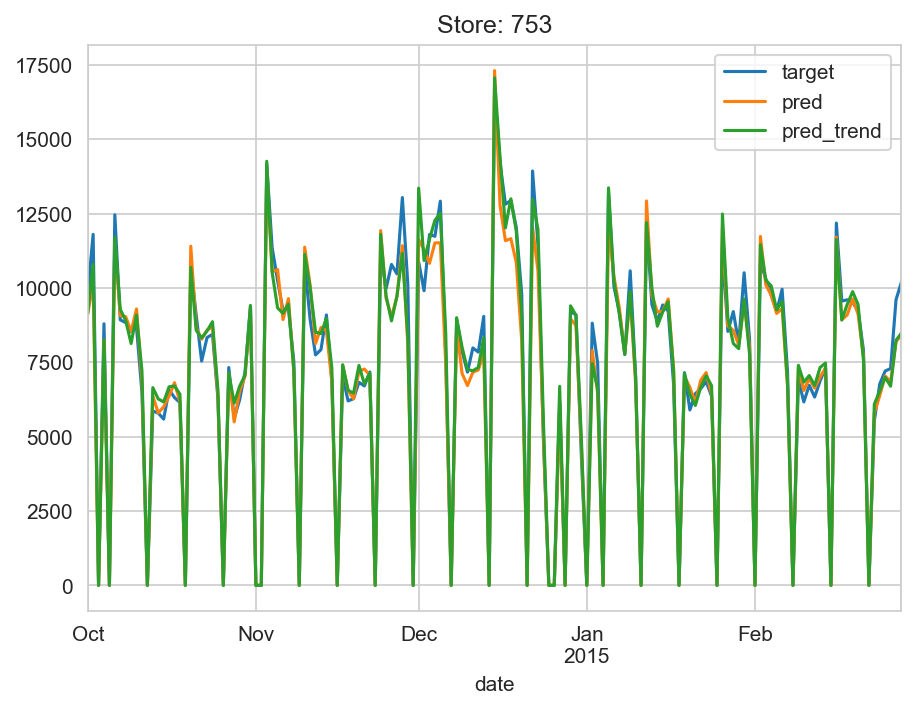}
 \caption{Store sales time series on validation dataset,  predicted  using model without (\textit{pred}) and with  (\textit{pred\_trend}) trend correction block ($RMSE=706, RMSE_{trend}=646$) }
 \label{img22}
 \end{figure}

\begin{figure}[H]
\center
 \includegraphics[width=0.65\linewidth]{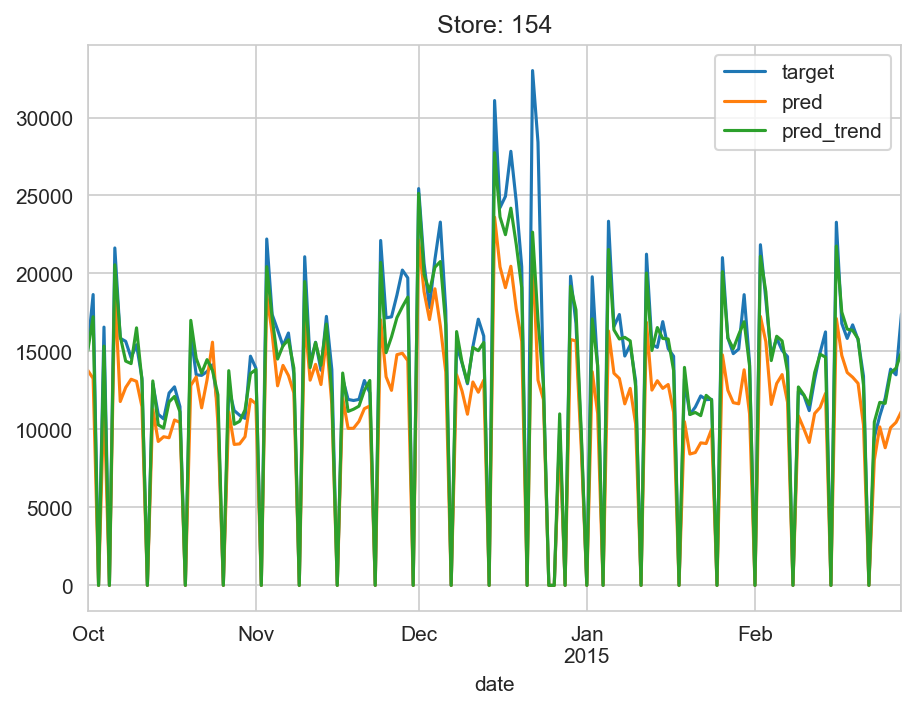}
 \caption{Store sales time series on validation dataset,  predicted  using model without (\textit{pred}) and with  (\textit{pred\_trend}) trend correction block ($RMSE=3699, RMSE_{trend}=1753$) }
 \label{img23}
 \end{figure}

\begin{figure}[H]
\center
 \includegraphics[width=0.65\linewidth]{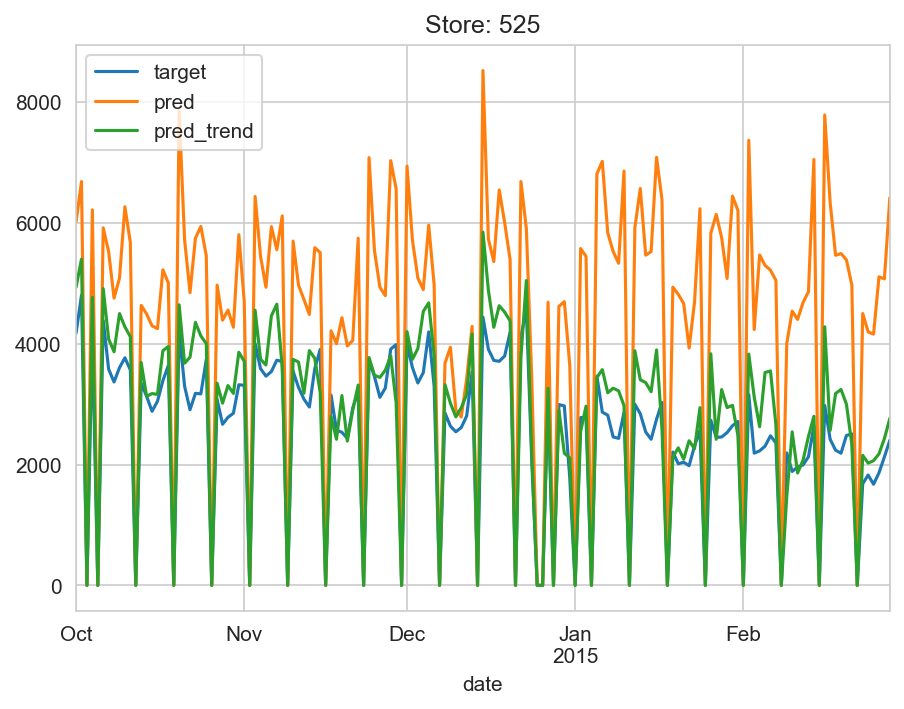}
 \caption{Store sales time series on validation dataset,  predicted  using model without (\textit{pred}) and with  (\textit{pred\_trend}) trend correction block ($RMSE=2515, RMSE_{trend}=572$) }
 \label{img24}
 \end{figure}

 Numerical features and target variables before feeding into the neural network were normalized by extracting their mean values from them and dividing them by their standard deviation. 
 The forecasting score over all stores is 
 $RMSE=1076$, for the model without the trend correction block and  $RMSE_{trend}=943$ for the model with the trend correction block.
 We can see a small improvement in the accuracy on the validation set for the model  with the trend correction block.  
 Figures~\ref{img22}-\ref{img24} show aggregated store sales time series on the validation dataset,  which was predicted  using the model without and with the  trend correction block.  These results of predicted sales correspond to time series which are  shown  in Figures~\ref{img1}-\ref{img3}.   One can see that  for some stores sales with time trend,  the forecasting accuracy  can be essentially improved using the trend correction block.  
 
\section{Conclusions}
Applying of machine learning for non-stationary sales time series with time trend can cause a forecasting bias.  
The approach with the trend correction block in the deep learning model for sales forecasting has been considered.  The model predicts sales values simultaneously with the weight for the time trend term.  The time trend term is considered as a product of the predicted weight value and normalized time value and then added to predicted sales values.  As a result, an optimized weight for the time trend for different groups of sales data can be received, e.g.  for each store with  the intrinsic time trend, the optimized weight for the time trend can be found. 
The results show that 
 the forecasting accuracy can be essentially improved for non-stationary sales with time trends using the trend correction block in the deep learning model. 

\bibliographystyle{ieeetr}
\bibliography{article.bib}

\end{document}